\documentclass{article}
\usepackage{ijcai19}

\usepackage{times}
\usepackage{soul}
\usepackage{url}
\usepackage[hidelinks]{hyperref}
\usepackage[utf8]{inputenc}
\usepackage[small]{caption}
\usepackage{mathrsfs}
\usepackage{graphicx}
\usepackage{amsmath}
\usepackage{amsthm}
\usepackage{booktabs}
\usepackage{multirow} 
\usepackage{amsfonts,amssymb}
\usepackage[linesnumbered,vlined,ruled]{algorithm2e}
\SetKwInOut{Input}{Input}\SetKwInOut{Output}{Output}
\newtheorem{definition}{Definition}

\usepackage{algpseudocode}

\newtheorem{lemma}{Lemma}

\newcommand{\mkp}{maximum $ k $-plex problem }
\newcommand{\rl}{reinforcement learning }

\title{Combining Reinforcement Learning and Configuration Checking for\\ Maximum k-plex Problem}

\author{
Peilin Chen$^1$\and
Hai Wan$^1$\and
Shaowei Cai$^{2}$\and
Weilin Luo$^1$\and
Jia Li$ ^1 $
\affiliations
$^1$School of Data and Computer Science, Sun Yat-sen University, Guangzhou, China\\
$^2$State Key Laboratory of Computer Science, Institute of Software, Chinese Academy of Sciences, China\\
\emails
chenpl7@mail2.sysu.edu.cn,
}

\begin{document}

\maketitle

\begin{abstract}
  The Maximum k-plex Problem is an important combinatorial optimization problem with increasingly wide applications. Due to its exponential time complexity, many heuristic methods have been proposed which can return a good-quality solution in a reasonable time. However, most of the heuristic algorithms are memoryless and unable to utilize the experience during the search. Inspired by the multi-armed bandit (MAB) problem in reinforcement learning (RL), we propose a novel perturbation mechanism named BLP, which can learn online to select a good vertex for perturbation when getting stuck in local optima. To our best of knowledge, this is the first attempt to combine local search with RL for the maximum $ k $-plex problem.
  Besides, we also propose a novel strategy, named Dynamic-threshold Configuration Checking (DTCC), which extends the original Configuration Checking (CC) strategy from two aspects.
  Based on the BLP and DTCC, we develop a local search algorithm named BDCC and improve it by a hyperheuristic strategy. The experimental result shows that our algorithms dominate on the standard DIMACS and BHOSLIB benchmarks and achieve state-of-the-art performance on massive graphs.
\end{abstract}

\vspace{-3mm}
\section{Introduction}

In social network analysis, detecting a large cohesive subgraph is a fundamental and extensively studied topic with various applications. 
Clique is a classical and ideal model in the field of cohesive subgraph detection. A graph is a clique if there is an edge between any pair of vertices. 
The Maximum Clique Problem, that is, to find a clique of maximum size in a given graph, is a fundamental problem in graph theory and finds wide application in many fields, such as biochemistry and genomics \cite{butenko2006clique}, wireless network \cite{lakhlef2015multi}, data mining \cite{boginski2006mining,conte2018d2k} and many others.

However, in some real-world applications, the networks of interest may be built based on empirical data with noises and faults.
In these cases, large cohensive subgraphs hardly appear as ideal clique. To tackle this problem, many clique relaxation models have been proposed. In this paper, we focus on $ k $-plex, a degree-based clique relaxation model. A simple undirect graph with $ n $ vertices is a $ k $-plex if each vertex of this graph has at least $ n - k $ neighbors. The \emph{maximum $ k $-plex problem}, that is, to find a $ k $-plex of maximum size on a given graph with a given integer $ k $, has received increasing attention from researchers in the fields of social network analysis and data mining \cite{xiao2017fast,conte2018d2k}.

The decision version of the maximum $ k $-plex problem is known to be NP-complete \cite{balasundaram2011clique}.
Different algorithms have been developed for this problem, including exact algorithms and heuristic ones.
\citeauthor{balasundaram2011clique} \shortcite{balasundaram2011clique} proposed a branch-and-bound algorithm based on a polyhedral study of this problem.
\citeauthor{mcclosky2012combinatorial} \shortcite{mcclosky2012combinatorial} developed two branch-and-bound algorithms adapted from combinatorial clique algorithms. 
Recently, \citeauthor{xiao2017fast} \shortcite{xiao2017fast} proposed an exact algorithm which breaks the trivial exponential bound of $ 2^{n} $ for \mkp with $ k\ge3 $. \citeauthor{gao2018exact} \shortcite {gao2018exact} proposed several graph reduction methods integrated them into a brand-and-bound algorithm.

Due to the exponential time complexity of the maximum $ k $-plex problem, several heuristic approaches have been proposed to provide a satisfactory solution within an acceptable time.
\citeauthor{DBLP:series/natosec/GujjulaSM14} \shortcite{DBLP:series/natosec/GujjulaSM14} proposed a hybrid metaheuristic based on the GRASP method. \citeauthor{miao2017approaches} \shortcite{miao2017approaches} improved the construction procedure to provide a better initial solution for GRASP method.
\citeauthor{zhou2017frequency} \shortcite{zhou2017frequency} developed a tabu search algorithm named FD-TS which achieved state-of-the-art performance. 


Local search is likened to ``trying to find the top of Mount Everest in a thick fog while suffering from amnesia'' \cite{russell2016artificial}. For a long time, much effort has been devoted to enable a memory mechanism for local search.
These works can be roughly divided into two parts. The first part focuses on exploiting the searching history to guide the search into a more promising area.
For example, \citeauthor{boyan2000learning} \shortcite{boyan2000learning} proposed the STAGE algorithm to learn an evaluation function from features of visited states which can be used to bias future search trajectory.
\cite{zhou2018improving} presented a probability learning based local search algorithm for the graph coloring problem.
The other part focuses on reducing the inherent cycling problem of local search. Tabu mechanism \cite{glover1998tabu} maintains a short-term memory of the recent search steps to forbid reversing the recent changes. Configuration Checking strategy \cite{cai2011local} keeps a memory of state change of local structures and reduces cycling problem by prohibiting cycling locally.
 
When getting stuck in local optima, a good perturbation mechanism can modify the candidate solution and generates a promising search area for the following search steps. Inspired by the multi-armed bandit problem and its algorithms, we propose the \emph{bandit learning based perturbation mechanism} (BLP), which can learn in an online way to select a good vertex for perturbation. To our best of knowledge, this is the first attempt to combine \rl and local search for the \mkp.

Recently, Configuration Checking (CC) and its variants have been successfully applied in various combinatorial optimization problems \cite{cai2011local,wang2016two,WangCCY18}, 
revealing the importance of exploiting the structural property of the problems. 
Different from tabu mechanism, CC is a non-parameter strategy which exploits the circumstance information to reduce cycling problem in local search.
However, CC and its variants have the following limitations. 
Firstly, the use of the configuration information is limited to handling the cycling problem. 
Secondly, the forbidding strength of the CC and its variants is static and cannot make adjustments to different problem instances.
In this paper, we propose a variant of CC, named Dynamic-threshold Configuration Checking (DTCC), to extend the original CC from two different aspects. One is the \emph{neighbor quality} heuristic which evaluates a vertex with consideration of the community it belongs to. The other is the \emph{dynamic threshold} mechanism which enables an adaptive forbidding strength for CC.

Based on BLP and DTCC, we develop a local search algorithm, called BDCC, and improve it by a hyperheuristic strategy. The resulting algorithm, named BDCC-H, can learn to select a good heuristic in adding and swapping phase. 
The experiments show that our algorithms dominate FD-TS on the standard DIMACS and BHOSLIB benchmarks. Not only is our algorithms robust and time-efficient, but also they provide better lower bounds on the size of the maximum $ k $-plexes for most hard instances. Besides, our algorithms achieve state-of-the-art performance on massive graphs.

The remainder of this paper is organized as follows. Section 2 gives some necessary background knowledge. Section 3 provides some formal definitions and proposes the BLP mechanism. Section 4 proposes DTCC strategy to extend CC from two aspects. In Section 5, we present the BDCC algorithm and improve it by a hyperheuristic strategy.
Section 6 shows the experimental results. Section 7 gives concluding remarks. 

\vspace{-3mm}
\section{Preliminaries}
\vspace{-1mm}
\subsection{Basic Definitions and Notations}\label{definition}
An undirected graph is defined as $ G=(V, E) $, where $ V $ is a set of vertices and $ E $ is a set of edges. Each edge $ e $ consists of two vertices, denoted as $e = (v, u) $, where $ v $ and $ u $ are the $ endpoints $ of this edge. Two vertices are \emph{neighbors} if they belong to an edge. Let $ N(v) $ denote the set of all neighbors of $ v $. The \emph{degree} of vertex $ v $ is defined as the $ deg(v) = |N(v)| $. For a vertex set $ S $, $ N(S)=\bigcup_{v \in S} N(v) \backslash S $ be the set of neighbors of $ S $ and $ G[S] = (S, E\cap(S\times S)) $ be the induced graph of $ S $. 

Given a graph $ G $ and an integer $ k\in$ $ \mathbb{Z^+} $, a subset $ S \subseteq V $ is a $ k $-plex, if $ |N(v) \cap S| \ge |S|-k $ for all $ v\in S $. A vertex $ v\in S $ is a \emph{saturated vertex} if $ |N(v) \cap S| = |S| - k $. The \emph{saturated set} $ \mathcal{C}[S] $ of set $ S $ is the set of all saturated vertices in $ S $.
 A vertex $ u\in S $ is {\em deficient vertex} if $ |N(u) \cap S| < |S| - k $. Obviously, any subset $ S\subseteq V $ containing deficient vertices cannot be a $ k $-plex.

A \emph{candidate solution} is a subset of $ V $. Given a graph $ G = (V, E) $, an integer $ k $ and a feasible $ k $-plex $ S \subseteq V $, 
a typical 3-phase local search algorithm for the maximum $ k $-plex problem 
 maintains a feasible $ k $-plex $ S $ as candidate solution and uses 
three operators, $ Add $, $ Swap $ and $ Perturb $ to modify it iteratively \cite{zhou2017frequency}. 
The set $ N(S) $ is split into three disjoint sets, $ AddSet(S) $, $ SwapSet(S) $ and $ PerturbSet(S) $, 
which contain the objects of the above-metioned operators. 
Here we give their formal definitions.

\vspace{-5mm}
\begin{align*}
AddSet(S) =& \{v\in N(S): |N(v) \cap S| > |S| - k \\
&\land \mathcal{C}[S]\backslash N(v) = \emptyset\} \\
SwapSet(S) = & \{ v\in N(S):\\
& |N(v) \cap S| \ge |S| - k \land |\mathcal{C}[S] \backslash N(v)| = 1) \\ 
&\lor |N(v) \cap S| = |S| - k \land |\mathcal{C}[S] \backslash N(v)| = \emptyset) \}\\
PerturbSet(S) =& N(S)\backslash (AddSet(S) \cup SwapSet(S)
\end{align*}

Obviously, the vertices in $ AddSet(S) $ can be added into $ S $ directly. The vertices in $ SwapSet(S) $ can be added into $ S $ while removing one vertex in $ S \backslash N(v) $. Adding a vertex $ v \in PerturbSet(S) $ into $ S $ would cause two or more deficient vertices in $ S \backslash N(v) $. Therefore, these vertices should be removed to maintain a feasible $ k $-plex.

\subsection{Multi-armed Bandit Problem}

\emph{Multi-armed Bandit Problem} (MAB) is a one-state RL problem. 
In this problem, there is a set of arms and an agent which repeatedly selects an arm to play at each step with a purpose to maximizing the long-term expected reward. 
Since the distribution of the reward of each arm is unknown, the agent faces an exploration-exploitation tradeoff. On the one hand, it needs to explore by selecting each arm to estimate the expected reward of them. On the other hand, it needs to exploit the existing knowledge by choosing arms with high expected rewards. 
The exploration-exploitation trade-off is a fundamental issue in reinforcement learning and the $ \epsilon $-greedy strategy is widely used to keep a balance of them. With the $ \epsilon $-greedy strategy, the agent chooses actions randomly for exploration with a probability $ \epsilon $ and makes choices greedily for exploitation with a probability $ 1 - \epsilon $ .

\vspace{1mm}
\subsection{Configuration Checking}
\emph{Configuration Checking} (CC) \cite{cai2011local}, is a parameter-free strategy that can exploit the structural property of the problem to reduce cycling problem in local search. The \emph{configuration} of a vertex is defined as the states of its neighbors.
The main idea of CC strategy is that if the configuration of a vertex remains unchanged since its last removal from candidate solution, then it is forbidden to be added back into the candidate solution.

Recently, different CC variants have been proposed and successfully applied to various combinatorial optimization problems \cite{cai2011local,DBLP:journals/jair/WangCY17,wang2017novel}. Here we highlight the Strong CC (SCC) strategy which was proposed in \cite{wang2016two} for the Maximum Weight Clique Problem. The difference between SCC and CC is that SCC allows a vertex $ v $ to be added into candidate solution only when some of $ v $'s neighbors have been added since $ v $'s last removal, while CC allows the adding of a vertex $ v $ when some of $ v $'s neighbors have been either added or removed.

Due to the similarity of clique and $ k $-plex, it is natural to think of applying SCC to local search for the maximum $ k $-plex problem. A straighforward SCC strategy for maximum $ k $-plex problem can be implemented as follows. We maintain a Boolean array $ confChange $ to indicate whether the configuration of each vertex has been changed. Only when a vertex $ v $ satisfies the SCC condition $ confChange(v) = 1 $ can it be added in $ S $. Initially we set $ confChange(v) = 1 $ for all $ v \in V $. When a vertex $ v $ is added into
candidate solution $ S $, for all $ v' \in N(v) $, $ confChange(v') $ is set to $ 1 $. When a vertex $ v $ is removed from $ S $,
$ confChange(v) $ is set to $ 0 $. As for a swap step, where vertex $ u $ is 
added into $ S $ at the cost of removal of vertex $ v $, $ confChange(v) $ is set to $ 0 $.

\vspace{-2mm}
\section{Learning from History: BLP Mechanism}

According to the definition in Subsection \ref{definition}, as $ S $ grows, $ AddSet(S) $ and $ SwapSet(S) $ become smaller because the vertices in them need to satisfy more constraints.
Therefore, $ PerturbSet(S) $ usually contains most of the vertices in $ N(S) $ when reaching a local optima. It is difficult to select a good vertex for perturbation from such a large set. 
We propose \emph{bandit learning based perturbation mechanism} (BLP) to learn from searching history to select a good vertex for perturbation in an online way.
In this section, we give some necessary formal definitions and present the BLP mechanism.


\begin{definition}
	Given a graph $ G = (V, E) $ and a $ k $-plex $ S \subseteq V $, an action is a pair $ (op, v) $ where $ op \in \{Add, Swap, Perturb\} $ is the operator and $ v \in V $ is the object. The available action set $ A(S) $ is defined as $ A(S) = \{Add\} \times AddSet(S) \cup \{Swap\} \times SwapSet(S) \cup \{Perturb\} \times PerturbSet(S) $. Let $ S \oplus a \rightarrow S' $ denote that applying action $ a \in A(S) $ to $ S $ results in a new $ k $-plex $ S' $.
\end{definition}

\begin{definition}
	A search trajectory is a finite sequence $ (S_0, S_1, \ldots, S_k) $ of $ k $-plexes $ S_i (i=0, 1, \ldots, k) $ such that for $ \forall i = 0, 1 \ldots, k - 1 $, $ \exists a_i \in A(S_i), S_i \oplus a_i \rightarrow S_{i + 1} $. 
\end{definition}

\begin{definition}
	The walk of a search trajectory $ \mathcal{T} = (S_0, S_1, \ldots, S_k) $ is an ordered action sequence $ \mathcal{A} = \left \langle a_0, a_1, \ldots, a_{k- 1} \right \rangle $ where $ a_i \in A(S_i) $ and $ S_i \oplus a_i \rightarrow S_{i + 1} $ for $ \forall i = 0, 1, \ldots, k - 1 $.
\end{definition}

\begin{definition}
	Given a search trajectory $ \mathcal{T} $, a $ k $-plex $ S_j $ in $ \mathcal{T} $ is a break-through point if $ |S_j| > |S_i| $ for $ \forall i = 0, 1, \ldots, j - 1 $, and an episode is a subsequence of $ \mathcal{T} $ between two adjacent break-through points. 
\end{definition}

The underlying consideration of BLP is that all the Perturb actions in the walk of an episode make contributions to the quality improvement at the end of this episode. 
The BLP treat each vertex as an arm in MAB and reward them according to their contribution when an episode is completed.
Therefore, the expected reward of a vertex can reflect the possibility of reaching another break-through point if perturbing the candidate solution with this vertex.
In the implementation, the BLP maintains a $ Q $ value for each vertex, initialized to 0 at the start of the search. When an episode is completed, we reward the objects of all the Perturb actions in the corresponding walk. For a vertex $ v $ to reward, we update $ Q(v) $ with \emph{exponential recency weighted average} (ERWA) technique \cite{sutton1998introduction}, as is shown in Equation \ref*{erwa}. 
 \begin{equation}
 \label{erwa}
 Q(v) = (1 - \alpha) \cdot Q(v) + \alpha \cdot r_v
 \end{equation}
 Here the $ \alpha \in [0,1] $ is a factor called \emph{stepsize} to determines the weight given to the recent reward, and $ r_v = \frac{1}{nPertb}$ where $ nPertb $ is the number of Perturb actions in the walk of this episode. The intuition behind the reciprocal reward value is that the actions applied in a shorter episode are more valuable than those in a longer one. In the perturbation phase, BLP selects a vertex with $ \epsilon $-greedy strategy.

\vspace{-3mm}
\section{Dynamic-threshold Configuration Checking}
According to our previous experiments, applying CC or other CC variants directly to the \mkp
does not lead to a good performance on graphs with high edge density.
The reason is that the configurations of the high-degree vertices in these graphs are very likely to change and CC (or other CC variants) cannot enhance its forbidding strength on these vertices.
To make better use of the configuration information and enable an adaptive forbidding strength, we propose a new variant of CC named Dynamic-threshold Configuration Checking (DTCC).
The two parts of DTCC are the \emph{neighbor quality} heuristic and \emph{dynamic threshold} mechanism. 

\subsection{Neighbor Quality Heuristic}
The \emph{neighbor quality} of a vertex $ v $, denoted by $ NQ(v) $, is defined as $ NQ(v) = \#N(v)_{in} - \#N(v)_{out} $, where $ \#N(v)_{in} $ (resp. $ \#N(v)_{out} $) is the total number of times a vertex in $ N(v) $ is added into (resp. removed from) the candidate solution. Due to the cohesive characteristics of $ k $-plex, a vertex that belongs to a higher-quality community is more likely to appear in a large $ k $-plex. In the implementation, DTCC maintains a integer $ NQ $ (initialized to $ 0 $) for each vertex, and update the $ NQ $ value with the following rule.

\textbf{DTCC-NQRule}. 
The $ NQ $ value is set to $ 0 $ for all $ v \in V $. 
When a vertex $ v $ is added into candidate solution $ S $, for all $ u \in N(v) $, $ NQ(u)++ $. When a vertex $ v $ is removed from $ S $, for all $ w \in N(v) $, $ NQ(w)-- $.

\vspace{-1.5mm}
\subsection{Dynamic Threshold Mechanism}
We extend the $ confChange $ to an integer array and maintain an integer array $ threshold $ that can adjust the forbidding strength on different vertices.
A vertex $ v $ is allowed to be added into candidate solution only when DTCC condition $ confChange(v) \ge threshold(v) $ is satisfied. The following four rules specify the dynamic threshold mechanism.

\textbf{DTCC-InitialRule}. In the beginning of search process, for all $ v \in V $, $ confChange(v) = threshold(v) = 1 $.

\textbf{DTCC-AddRule}. When $ v $ is added into candidate solution, $ confChange(v) = 0 $, $ threshold(v)++ $, and for all $ v' \in N(v) $, $ confChange(v')++ $.

\textbf{DTCC-SwapRule}. When $ v $ is added into candidate solution at the cost of removal of $ u $, $ confChange(u) = 0 $.

\textbf{DTCC-PerturbRule}. When $ v $ is added into candidate solution and a set of vertices $ X \subseteq S \backslash N(v) $ is removed from this candidate solution, $ confChange(w) = 0 $ for all $ w \in X $, $ confChange(v) = 0 $, $ threshold(v)++ $ and $ confChange(v')++ $ for all $ v' \in N(v) $.

Note that the SCC strategy is a special case of DTCC strategy whose $ confChange $ is a Boolean array and $ threshold $ is fixed to $ 1 $. Lemma \ref{relation} illustrate their relation.

\begin{lemma}\label{relation}
	If a vertex $ v $ satisfies the DTCC condition, then it satisfies the SCC condition. The reverse is not necessarily true.
\end{lemma}
\vspace{-3mm}
\begin{proof}
	According to DTCC rules, $ threshold(v) \ge 1 $ for $ \forall v \in V $ during the search processs. If the the DTCC condition $ confChange(v) \ge threshold(v) $ holds, then $ confChange(v) \ge threshold(v) \ge 1 $. So at least one neighbor of $ v $ must be added into candidate solution since the last time $ v $ was removed. So the SCC condition is satisfied.
	
	Suppose $ v $ satisfies the SCC condition $ confChange(v) = 1 $, but $ threshold(v) > confChange(v) $. In this case the DTCC condition is not satisfied.
\end{proof}
\vspace{-2mm}

According to Lemma \ref{relation}, we can conclude that DTCC has stronger forbidding strength than SCC. Generally, a frequently operated vertex has a high $ threshold $ and is more likely to be forbidden. Thus the algorithm is forced to select other vertices to explore the search space. 


\vspace{-2mm}
\section{BDCC Algorithm and A Hyperheuristic}

\subsection{BDCC Algorithm}
Based on the BLP mechanism and DTCC strategy, we develop a local search algorithm named BDCC, whose pseudocode is shown in Algorithm \ref{slpdtcc}.
Initially, the best found $ k $-plex, denoted as $ S^* $, is initialized as empty set. 
In each loop (line 3-10), an initial solution is firstly constructed (line 4) as the starting point of the search trajectory, and the search procedure starts. If the best solution in this search trajectory $ S_{lbest} $ is better than the best solution ever found $ S^* $, $ S^* $ is updated by $ S_{lbest} $ and $ Peel() $ function (line 8) is called to reduce the graph. If the reduced graph has fewer vertices than $ |S^*| $, then $ |S^*| $ is returned as one of the optimum solutions. Three major components in BDCC are initial solution construction, search procedure and graph peeling. We describe them in detail in the following.


\begin{algorithm}
	\caption{BDCC algorithm}
	\label{slpdtcc}
	\Input {A graph $ G $, an integer $ k $, time limit $ cutoff $, iterations limit $ L $}
	\Output {The largest $ k $-plex found}
		 $ S^* \leftarrow \emptyset $\;
		 $ NQ(v) \leftarrow 0 $, $ opTimes(v) \leftarrow 0 $, $ Q(v) \leftarrow 0 $ for all $ v $\;
		\While {$ elapsed time < cutoff $}{
			 $ S \leftarrow \textsc{ConstructInitSolution}(G, k, opTimes) $\;
			 $ S_{lbest} \leftarrow \textsc{Search}(G, k, S, NQ, opTimes, Q, L) $\;
			 
	

			 \If {$ |S_{lbest}| >|S^*| $}{
				 $ S^* \leftarrow S_{lbest} $\;
				 $ G \leftarrow \textsc{Peel}(G, k, |S^*|) $	\;
			 }
			 
			 \If {$ |V| \leq |S^*| $}{
				 \textbf{return} $ S^* $\;
			 }
	
		}
		\textbf{return} $ S^* $

\vspace{-1mm}
\end{algorithm}


We adopt the construction function in FD-TS \cite{zhou2017frequency}. The $ \textsc{ConstructInitSolution}() $ function firstly use the vertex with minimum $ opTimes $ (breaking ties randomly) in a random sample of $ 100 $ vertices to create the singleton set $ S $, and repeatedly add the vertex with minimum $ opTimes $ (breaking ties randomly) in $ AddSet(S) $ into $ S $ until $ AddSet(S) $ is empty. Then the final $ k $-plex is returned as the initial solution. By giving priority to vertices that are operated less frequently, the construction procedure can generate diversified initial solutions in different rounds.

\begin{algorithm}[h]
	\caption{\textsc{Search}($ G, k, S, NQ, opTimes, Q, L $)}
	\label{search}
	\Input {A graph $ G $, an integer $ k $, initial solution $ S $, integer array $ NQ $, integer array $ opTimes $, floatint point number array $ Q $, search depth $ L $}
	\Output {The largest $ k $-plex in this search procedure}
	
	$ S_{lbest} \leftarrow S $, $ walk \leftarrow \left \langle \right \rangle $\;
	$ confChange(v) = threshold(v)=1 $ for all $  v \in V $\;
	\For {$ curStep = 0 $; $ curStep < L $; $ curStep++ $}{
		split $ \{v \in N(S) \mid v $ satisfies DTCC condition $\} $ into $ AddSet(S), SwapSet(S) $ and $ PerturbSet(S) $\;
		
		\If {$ AddSet(S) \neq \emptyset $}{
			$ v \leftarrow $ $ v \in AddSet(S) $ with biggest $ NQ(v) $, breaking ties randomly\;

			$ op \leftarrow Add $\;
		}

		\ElseIf{$ SwapSet(S) \neq \emptyset $}{
			$ v \leftarrow $ $ v \in SwapSet(S) $ with biggest $ NQ(v) $, breaking ties randomly\;
			$ op \leftarrow Swap $\;
		}

		\Else{
			\If {$ PerturbSet(S) \neq \emptyset $}{
				$ v \leftarrow $ select $ v $ with $ \epsilon$-greedy mothod\;
				$ op \leftarrow Perturb $\;
				
			}
			\Else{
				\textbf{return} $ S_{lbest} $\;
			}
			
		}
		$ S \leftarrow S \oplus (op, v) $, $ opTimes(v)++ $\;
		$ walk \leftarrow walk \cdot \left \langle (op, v) \right \rangle $\;
		update $ NQ $, $ confChange $ and $ threshold $ according to the DTCC rules\;
		
		\If {$ S > |S_{lbest}| $}{
			$ S_{lbest} \leftarrow S $\;
			reward the Perturb actions in $ walk $\;
			$ walk \leftarrow \left \langle \right \rangle $\;
		}
	}
	\textbf{return} $ S_{lbest} $\;
\end{algorithm}

The $ \textsc{Search}() $ function iteratively selects one action to modify the candidate solution until the iterations limit $ L $ is reached, or the available action set is empty, as is shown in Algorithm \ref{search}. The $ S_{lbest} $ records the best-quality solution in the search trajectory so far, and the $ walk $ record the action sequence since the last break-through point. 
The algorithm selects vertices with the highest $ NQ $ for adding and swapping and selects vertices for perturbation according to their $ Q $ with $ \epsilon $-greedy strategy.
After each iteration, if $ |S| > |S_{lbest}| $, that means a break-through point is reached and the episode is completed, then the objects of all Perturb actions in current $ walk $ (if there exist) will be rewarded with the ERWA algorithm and $ walk $ will be cleared out.


If the $ S_{lbest} $ returned by $ \textsc{Search}() $ is better than $ S^* $, then $ S^* $ is updated with $ S_{lbest} $ and the $ \textsc{Peel}() $ function is called to recursively deletes the vertices (and their incident edges) with a degree less than $ |S^*| - k + 1 $ until no such vertex exists. It is sound to remove these vertices since they can not be included in any feasible $ k $-plexes larger than $ |S^*| $.

\subsection{Improving BDCC by A Hyperheuristic}

Our previous experiments show that selecting vertices from $ AddSet(S) $ and $ SwapSet(S) $ greedily can usually lead to a high-quality solution on most hard instances. However, on some problem domains where the optimum solutions are hidden by incorporating low-degree vertices, the search may be misled by the greedy manner and miss the best solutions in some runs \cite{cai2011local}. To enhance the robustness of BDCC, we design a hyperheuristic based on simulated annealing to switch between different heuristics dynamically and select the suitable one for different problem instances. We equip BDCC with this hyperheuristic, developing an algorithm named BDCC-H, as outlined in Algorithm \ref{slpdtcchl}.

\begin{algorithm}
	\caption{BDCC-H Algorithm}
	\label{slpdtcchl}
	\Input {A graph $ G $, an integer $ k $, time limit $ cutoff $, search depth $ L $, initial temperature $ T $, cooling rate $ \gamma $}
	
	\Output {The largest $ k $-plex found}
	$ S^* \leftarrow \emptyset $, $ best_{H_i} \leftarrow 0 $ for each $ i $\;

	\While {$ elapsed time < cutoff $}{
		$ H_{selected} \leftarrow $ select a heuristics under current $ T $\;	
		Construct a initial solution $ S $\;
		$ S_{lbest} \leftarrow $ search with $ H_{selected} $\;
		\If {$ best_{H_{selected}} < |S_{lbest}| $}{
			$ best_{H_{selected}} = |S_{lbest}| $\;
		}
		\If {$ |S^*| < |S_{lbest}| $}{
			$ S^* < S_{lbest} $\;
			$ G \leftarrow \textsc{Peel}(G, k, |S^*|) $\;
		}
		\If {$ T > 0.01 $}{
			$ T \leftarrow \gamma \cdot T $\;
		}
	}
	\vspace{-2mm}
\end{algorithm}

The difference between BDCC and BDCC-H is whether the heuristics for adding and swapping is fixed. The BDCC-H algorithm adopts three heuristics for adding and swapping, (i)$ H_1 $, selecting vertex with largest $ NQ $, (ii)$ H_2 $, selecting vertex with largest $ Q $, (iii)$ H_3 $, selecting vertex randomly. The BDCC-H maintains a variable $ best_{H_i} $ for each heuristic $ H_i $ to record the size of the best solution found with $ H_i $. A temperature $ T $ is used to control the heuristic selection. 
Before the search procedure begins, the algorithm selects one heuristic under current $ T $. 
The selection probability of each heuristic $ H_i $ is defined based on Boltzmann distribution $ \frac{e^{best_{H_i} / T}}{\sum_{j=1}^{3} e^{best_{H_j} / T} } $, which is widely used for softmax selection \cite{sutton1998introduction}. A relatively high initial temperature can lead to equal selection to force exploration. As the temperature $ T $ cools down, the algorithm are more inclined to select a heuristic with highest $ best $ value and exploit with this heuristic.

%
%

\vspace{-2mm}
\section{Experimental Result}

We evaluate our algorithms on standard DIMACS and BHOSLIB benchmarks as well as massive real-world graphs.
\begin{table*}[h]
	\centering
	\caption{Experimental Result on DIMACS with $ k = 2, 3, 4 $}
	\vspace{-3mm}
	\tiny{                 
		\setlength{\tabcolsep}{1mm}{      
			\begin{tabular}{l|cccc|cccc|cccc}
				\toprule
				\multirow{2}{*}{Instance}
				& \multicolumn{4}{|c|}{k=2}& \multicolumn{4}{|c|}{k=3}& \multicolumn{4}{|c}{k=4} \\
				\cline {2 -13}
				& FD-TS & BDCC & BDCC-H & $\Delta _{time}$  & FD-TS & BDCC & BDCC-H & $\Delta_{time}$& FD-TS & BDCC & BDCC-H & $\Delta_{time}$ \\
				\midrule
				brock400\_4 & 33(\textbf{33}) & 33(32.88) & 33(\textbf{33}) & \bfseries-236.95 & 36(36) & 36(36) & 36(36) & \bfseries-1.25 & 41(41) & 41(41) & 41(41) & \bfseries-0.25 \\
				brock800\_1 & 25(25) & 25(25) & 25(25) & \bfseries-1.79 & 30(29.92) & 30(29.34) & 30(29.92) & \bfseries-69.77 & 34(34) & 34(33.96) & 34(34) & \bfseries-35.45 \\
				brock800\_2 & 25(25) & 25(25) & 25(25) & \bfseries-1.01 & 30(30) & 30(29.98) & 30(30) &\bfseries -159.13 & 34(33.96) & 34(33.36) & 34({\bfseries34}) &  \\
				brock800\_3 & 25(25) & 25(25) & 25(25) & \bfseries-1.59 & 30(30) & 30(29.64) & 30(30) & \bfseries-39.76 & 34(34) & 34(33.6) & 34(34) & \bfseries-28.83 \\
				brock800\_4 & 26(26) & 26(25.78) & 26(26) &\bfseries -4.59 & 29(29) & 29(29) & 29(29) & \bfseries-3.83 & 34(33.12) & 34(33.02) & 34({\bfseries33.22}) &  \\
				C1000.9 & 82(81.56) & 82(81.9) & 82({\bfseries82}) &  & 96(95.14) & 96({\bfseries95.32}) & 96(95.22) &  & 109(107.62) & \textbf{110}({\bfseries108.32}) &\textbf{110}(108.14) &  \\
				C2000.5 & 20(19.6) & 20(19.86) & 20({\bfseries19.94}) &  & 23(22.14) & 23(22.18) & 23({\bfseries22.4}) &  & 26(25.04) & 26(\bfseries25.06) & 26(25.04) & \bfseries-38.25 \\
				C2000.9 & 92(90.7) & \bfseries94(92.44) & 93(91.98) &  & 106(105.14) & \bfseries109(107.22) & 108(107.02) &  & 120(118.6) & \bfseries123(121.14) & \textbf{123}(121.02) &  \\
				C4000.5 & 21(20.5) & \bfseries22(21.02) & 21(20.92) &  & 24(23.38) & 24({\bfseries24}) & 24(23.9) &  & 27(26.12) & \bfseries28(27) & 27(26.74) & \\
				DSJC1000.5 & 18(18) & 18(18) & 18(18) & \bfseries-2.96 & 21(21) & 21(21) & 21(21) &\bfseries -3.64 & 24(23.98) & 24(23.56) & 24({\bfseries24}) &  \\
				gen400\_p0.9\_65 & 74(72.68) & 74({\bfseries73.22}) & 74(73.14) &  & 101(100.96) & 101(101) & 101({\bfseries101}) &  & 132(132) & 132(132) & 132(132) & \bfseries-0.02 \\
				gen400\_p0.9\_75 & 79(78.74) & 79(79) & \bfseries80(79.02) &  & 114(114) & 114(114) & 114(114) & \bfseries-0.01 & 136({\bfseries136}) & 136(131.8) & 136(132.22) & \\
				keller5 & 31(31) & 31(31) & 31(31) &\bfseries -0.01 & 45(45) & 45(45) & 45(45) & \bfseries-2.26 & 53(53) & 53(53) & 53(53) & \bfseries-3.64 \\
				keller6 & 63(63) & 63(63) & 63(63) & \bfseries-0.50 & 93({\bfseries90.28}) & 93(90.06) & 93(90.12) &  & 109(106.28) & 113({\bfseries109.3}) & {\bfseries117}(108.9) & \\
				MANN\_a45 & 662({\bfseries661.28}) & 661(661) & 662(661.02) &  & 990(990) & 990(990) & 990(990) & \bfseries-1.83 & 990(990) & 990(990) & 990(990) &\bfseries -1.94 \\
				MANN\_a81 & 2162({\bfseries2161.34}) & 2162(2161.04) & 2162(2161.08) &  & 3240(3240) & 3240(3240) & 3240(3240)& \bfseries-101.04 & 3240(3240) & 3240(3240) & 3240(3240) & \bfseries-105.40 \\
				p\_hat1500-2 & 80(80) & 80(80) & 80(80) & \bfseries-0.01 & 93(93) & 93(93) & 93(93) & \bfseries-0.04 & 107(\textbf{107}) &  107(106.94) & 107(\textbf{107}) & \bfseries-59.27 \\
				san400\_0.7\_2 & 32(31.98) & 32(31.98) & 32({\bfseries32}) &  & 47(46.04) & 47(46.04) & 47({\bfseries46.68}) &  & 61(61) & 61(61)  & 61(61) & +0.02 \\
				san400\_0.7\_3 & 27(26.4) & 27({\bfseries27}) & 27({\bfseries27}) &  & 38(37.96) & 39(38.04) & \bfseries39(38.04) &  & 50(49.12) & 50(49.46) & 50({\bfseries50}) & \\
				san400\_0.9\_1 & 102(101.36) & 103(102.14) & \bfseries103(102.24) &  & 150(150) & 150(150) & 150(150) &\bfseries -0.02 & 200(200) & 200(200) & 200(200) & \bfseries-0.02 \\
				\bottomrule
			\end{tabular}
		}
	}
	\vspace{-2mm}
	\label{dimacs2}
\end{table*}
\vspace{-2mm}

\begin{table*}[h]
	\centering
	\caption{Experimental Result on BHOSLIB with $ k = 2, 3, 4 $}
	\vspace{-3mm}
	\tiny{                 
		\setlength{\tabcolsep}{2mm}{      
			\begin{tabular}{l|ccc|ccc|ccc}
				\toprule
				\multirow{2}{*}{Instance}
				& \multicolumn{3}{|c|}{k=2}& \multicolumn{3}{|c|}{k=3}& \multicolumn{3}{|c}{k=4} \\
				\cline {2 - 4}\cline{5 - 7}\cline{8 - 10}
				& FD-TS & BDCC & BDCC-H  & FD-TS & BDCC & BDCC-H & FD-TS & BDCC & BDCC-H \\
				\midrule
				frb50-23-1  & 67(66.2)    & 67({\bfseries66.28}) & 67(66.14) & 79(78.26) & 79(78.92) & 79({\bfseries78.98}) & 92(90.3)   & 92(91.12) & 92({\bfseries91.36}) \\
				frb50-23-2  & {\bfseries67}(66) 	  & 66(66)    & 66(66)    & 79(78.22) & 79(78.88) & 79({\bfseries79})    & 91(90.04)  & 91(90.88) & \bfseries92(90.98) \\
				frb50-23-3  & 65(63.96)   & 65({\bfseries64.06}) & 65(64.04) & 76(75.32) & 76({\bfseries75.98}) & \bfseries77(75.98) & 87(86.36)  & {\bfseries88}(87.52) & \bfseries88(87.68) \\
				frb50-23-4  & 66(65.56)   & 66(65.94) & 66({\bfseries65.96}) & 79(77.88) & 79(78.66) & 79({\bfseries79})    & 91(89.62)  & {\bfseries92}(90.76) & 91({\bfseries91}) \\
				frb50-23-5  & 67(66.1)    & 67({\bfseries66.22}) & 67(66.14) & 79(78.14) & \bfseries80(79)    & {\bfseries80}(78.88) & 91(90.14)  & \bfseries92(91.18) & {\bfseries92}(90.98) \\
				frb53-24-1  & 71(69.52)   & 71({\bfseries70.8})  & 71(70.48) & 85(83.24) & 85({\bfseries84.24}) & 85(84)    & 97(96.38)  & \bfseries99(97.74) & 98(97.56) \\
				frb53-24-2  & 70(69.02)   & 70(69.62) & 70({\bfseries69.98}) & 83(81.26) & 83(82.12) & 83({\bfseries82.2})  & 94(93.34)  & \bfseries96(94.82) & 95({\bfseries94.82}) \\
				frb53-24-3  & 70(69.16)   & 70(69.74) & 70({\bfseries69.82}) & 83(82.06) & 83({\bfseries82.94}) & 83({\bfseries82.94}) & 96(94.78)  & {\bfseries97}(96.22) & \bfseries97(96.32) \\
				frb53-24-4  & 70(68.64)   & 70({\bfseries69.08}) & 70(68.96) & 82(81.66) & {\bfseries84}(82.76) & \bfseries84(82.88) & 96(94.22)  & 96(95.72) & 96({\bfseries95.88}) \\
				frb53-24-5  & 68(67.72)   & 68({\bfseries68})    & 68({\bfseries68}) 	  & 82(80.16) & 82({\bfseries81.34}) & 82(81.18) & 93(92.18)  & \bfseries94(93.48) & {\bfseries94}(93.3) \\
				frb56-25-1  & 75(73.5)    & 75({\bfseries74.12}) & 75(74.08) & 89(87.78) & 89({\bfseries88.88}) & 89(88.84) & 103(101.4) & {\bfseries104}(102.92) & \bfseries104(103.02) \\
				frb56-25-2  & 74(73.42)   & \bfseries75(74.46) & {\bfseries75}(74.34) & 88(87.14) & \bfseries89(88.54) & {\bfseries89}(88.44) & 102(100.36)& \bfseries103(101.96) & 102(101.56) \\
				frb56-25-3  & 74(72.58)   & 74({\bfseries73.3})  & 74(73.18) & 87(85.72) & {\bfseries88}(87.16) & \bfseries88(87.62) & 100(98.66) & \bfseries101(100.02) & {\bfseries101}(99.94) \\
				frb56-25-4  & 73(72.28)   & \bfseries74(73.12) & {\bfseries74}(72.88) & 87(85.2)  & \bfseries88(86.94) & {\bfseries88}(86.44) & 99(98.16)  & \bfseries101(99.66)  & 100(99.32) \\
				frb56-25-5  & 74(72.48)   & 74({\bfseries73.04}) & 74(72.9)  & 87(85.74) & \bfseries88(86.92) & 87(86.74) & 100(98.68) & \bfseries101(100.12) & {\bfseries101}(100.08) \\
				frb59-26-1  & 78(77.14)   & \bfseries79(78.04) & {\bfseries79}(78.02) & 92(91.04) & {\bfseries93}(92.76) & \bfseries93(92.92) & 106(105.1) & \bfseries108(107.06) & {\bfseries107}(106.92) \\
				frb59-26-2  & 78(77.16)   & \bfseries79(78.08) & 78(77.92) & {\bfseries94}(91.34) & \bfseries94(92.88) & 93(92.78) & 106(105.18)& \bfseries108(106.84) & 107(106.6) \\
				frb59-26-3  & 77(76.04)   & \bfseries78(76.84) & 77(76.56) & 92(90.38) & \bfseries94(91.72) & 92(91.1)  & 106(104.24)& \bfseries107(105.64) & 106(105.26) \\
				frb59-26-4  & 77(76.18)   & \bfseries78(77.12) & 77(76.94) & 91(90.24) & \bfseries93(92)    & {\bfseries93}(91.8)  & 105(104.02)& \bfseries107(105.78) & 106(105.6) \\
				frb59-26-5  & 78(76.24)   & 78({\bfseries77.62}) & 78(77.52) & 92(90.08) & \bfseries93(91.94) & 92(91.74) & 104(103.38)& \bfseries107(105.62) & 106(105.5) \\
				frb100-40   & 126(123.82) & \bfseries127(124.42) & 126(124) & 149(146.86) & 149(147) & 149({\bfseries147.38}) & 170(168.64) & \bfseries173(169.88) & 170(169.14) \\
				
				\bottomrule
			\end{tabular}
		}
	}
	\vspace{-3mm}
	\label{bhoslib}
\end{table*}

\subsection{Experiment Preliminaries}
BDCC and BDCC-H and their competitor FD-TS are all implemented in C++ and compiled by g++ with '-O3' option. All experiments are run on an Intel Xeon CPU E7-4830 v3 @ 2.10GHz with 128 GB RAM server under Ubuntu 16.04.5 LTS. We set the search depth $ L = $  $ 1000 $, the stepsize $ \alpha = 0.5 $ and $ \epsilon = 0.2 $ for BDCC and BDCC-H. The initial $ T $ and cooling rate $ \gamma $ are set to $ 1000 $ and $ 0.99 $ respectively for BDCC-H. The cutoff time of each instance is set to $ 1000 $ seconds. All algorithms are executed 50 independently times with different random seeds on each instance with $ k = 2,3,4 $. 

\vspace{-1mm}
\subsection{Evaluation on DIMACS and BHOSLIB}
\vspace{-1mm}
We carried out experiments on standard DIMACS and BHOSLIB benchmark to evaluate our algorithms. The DIMACS benchmark taken from the Second DIMACS Implementation Challenge \cite{johnson1996cliques} includes problems from the real world and randomly generated graphs. The BHOSLIB instances are generated randomly based on the model RB in the phase transition area \cite{xu2005simple} and famous for their hardness. 

Table \ref{dimacs2} and Table \ref{bhoslib} show the experimental results on these two benchmarks. We report the best size and average size of $ k $-plex found by our FD-TS and algorithms, and compare the average time cost of BDCC-H and FD-TS if they have the same best and average solution sizes, shown in column $ \Delta_{time} $.
Most DIMACS instances are so easy that all the three algorithms find the same-quality solution very quickly, and thus are not reported. The result shows that our algorithm not only finds $ k $-plexes that FD-TS cannot reach on many instances but usually cost less time than FD-TS on other instances. Particularly, BDCC dominates on the CXXXX.X domain but is not robust enough on the brock domain. With a hyperheuristic, BDCC-H enhances the robustness of BDCC while achieving better performance than FD-TS. Remark that for C1000.9 with $ k=2 $, san400\_0.7.3 with $ k =4 $ and DSJC1000.5 with $ k = 4 $, BDCC-H is the only algorithm that find $ k $-plexes of size $ 82, 50 $ and $ 24 $ respectively in $ 100\% $ runs. 

On the BHOSLIB benchmark, BDCC and BDCC-H dominate FD-TS on most of the instances. We highlight the frb100-40, the hard challenging instance in BHOSLIB. BDCC updates the lower bound of the size of maximum $ 2 $-plex and $ 3 $-plex on frb100-40, indicating its power on large dense graphs.
Though BDCC-H does not achieve the same performance as BDCC due to the time-consuming exploration phase of hyperheuristic, it outperforms FD-TS on most of these instances.
%
\vspace{-1mm}
\subsection{Evaluation on Massive Graphs}
\vspace{-1mm}
We also evaluate our algorithm on massive real-world graphs from Network Data Repository \cite{DBLP:conf/aaai/RossiA15}, 
Thanks to the powerful peeling technique, most of these graphs are reduced significantly and solved in a short time by BDCC and BDCC-H. For other instances, our algorithms and FD-TS find a solution of the same quality in $ 100\% $ runs. So we do not report the results. To further assess performance on massive graphs, we choose the state-of-the-art exact algorithm named BnB \cite{gao2018exact} for comparison. We run BnB with a cutoff time of 10000 seconds for the optimum solution. For the sake of space, we do not report the instances that can be solved by both BnB and BDCC-H in a few seconds. Table \ref{massivegraph} shows the best solution found by BDCC-H and the average time cost of BnB and BDCC-H. An item with a symbol ``$ * $''  in column ``max'' indicates that this is the size of the optimum solution proved by BnB.

The result in Table \ref{massivegraph} shows that BDCC-H can find an optimum solution on most instances while costing much less time. For the instances that BnB fails to solve in 10000 seconds, BDCC-H can return a satisfactory solution within a few seconds.

\begin{table}[h]
	\centering
	\caption{Experimental Result on Massive Graphs with $ k = 2, 4 $}
	\vspace{-3mm}
	\tiny{                 
		\setlength{\tabcolsep}{0.2mm}{      
			\begin{tabular}{l|c|c|ccc|ccc}
				\hline
				\multirow{2}{*}{Instance}
				& &  & \multicolumn{3}{|c|}{k=2}& \multicolumn{3}{|c}{k=4} \\
				\cline {4 -9}
				& V & E & max & BnB & BDCC-H & max & BnB & BDCC-H \\
				\hline
				ca-coauthors-dblp.clq	& 540486		& 15245729	& $337^{*}$ & 21.132	& \bfseries0.8642	& $337^{*}$ & 21.392	& \bfseries0.6715  \\
				ia-wiki-Talk.clq	& 92117			& 360767	& $18^{*}$& 14.756	& \bfseries0.0379	& $23^{*}$& 775.832	& \bfseries0.0822 \\
				inf-road-usa.clq	& 23947347		& 28854312	& $5^{*}$& 50.768		& \bfseries6.5526	& 7	& $ > $10000		& \bfseries5.8993 \\
				inf-roadNet-CA.clq	& 1957027		& 2760388	& $5^{*}$& 3.78		& \bfseries0.5032	& 7	& $ > $10000		& \bfseries0.4594 \\
				inf-roadNet-PA.clq	& 1087562		& 1541514	& $5^{*}$& 1.448		& \bfseries0.2505	& 7	& $ > $10000		& \bfseries0.2486 \\
				sc-nasasrb.clq	        & 54870			& 1311227	& $24^{*}$& 820.244	& \bfseries0.0204	& 24	& $ > $10000		& \bfseries0.0098 \\
				sc-pkustk11.clq	        & 87804			& 2565054	& $36^{*}$& 7.348		& \bfseries3.6641	& $36^{*}$& 74.808	& \bfseries7.7031 \\
				sc-pkustk13.clq	        & 94893			& 3260967	& $36^{*}$& 560.98	& \bfseries0.1012	& 36	& $ > $10000		& \bfseries0.0792 \\
				sc-shipsec1.clq	        & 140385		& 1707759	& $24^{*}$& 1.952		& \bfseries0.3167	& $24^{*}$& 9311.244	& \bfseries0.3665 \\
				sc-shipsec5.clq	        & 179104		& 2200076	& $24^{*}$& 38.372	& \bfseries0.0988	& 24	& $ > $10000		& \bfseries0.1255 \\
				socfb-A-anon.clq	& 3097165		& 23667394	& $28^{*}$& 208.696	& \bfseries32.796	& $35^{*}$& 501.744	& \bfseries52.559 \\
				socfb-B-anon.clq	& 2937612		& 20959854	& $27^{*}$& 1128.236	& \bfseries21.414	& 33	& $ > $10000		& \bfseries22.911 \\
				tech-as-skitter.clq	& 1694616		& 11094209	& $69^{*}$& 3058.656	& \bfseries1.42		& $74^{*}$& 1829.164	& \bfseries1.746 \\
				tech-RL-caida.clq	& 190914		& 607610	& $20^{*}$& 2.272		& \bfseries0.103	& $24^{*}$& 1371.484	& \bfseries0.138 \\
				web-it-2004.clq	        & 509338		& 7178413	& $432^{*}$& 15.22	& \bfseries0.437	& $432^{*}$& 355.148	& \bfseries0.49 \\
				web-uk-2005.clq		& 129632		& 11744049	& $500^{*}$& 18.484	& \bfseries0.484	& $500^{*}$& 525.108	& \bfseries0.459 \\
				web-wikipedia2009.clq	& 1864433		& 4507315	& $32^{*}$& 185.24	& \bfseries1.06		& 32	 & $ > $10000	& \bfseries0.828 \\
				
				\hline
			\end{tabular}
		}
	}
	
	\label{massivegraph}
\end{table}

\vspace{-2mm}
\section{Conclusions and Futrue Work}

In this paper, we have proposed two heuristics, BLP and DTCC, for the maximum $ k $-plex problem. Based on BLP and DTCC, we develop a local search algorithm BDCC and further improve it by applying a hyperheuristic strategy for the adding and swapping phase. The experimental result shows that our algorithms achieve high robustness across a broad range of problem instances and update the lower bounds on the size of the maximum $ k $-plexes on many hard instances. Meanwhile, our algorithm achieve state-of-the-art performance on massive real-world.
\vspace{0mm}

In the future, we plan to study variants of CC for other combinatorial optimization problems further. Besides, it would be interesting to adapt the ideas in this paper to design local search algorithms for other clique relaxation model.

\bibliographystyle{named}
\bibliography{ijcai19}

\end{document}